\documentclass[11pt]{article}

\usepackage[preprint]{acl}

\usepackage{times}
\usepackage{latexsym}

\usepackage[T1]{fontenc}

\usepackage[utf8]{inputenc}

\usepackage{microtype}

\usepackage{inconsolata}

\usepackage{graphicx}

\usepackage{tabularx} 
\usepackage{amssymb} 
\usepackage{hyperref}
\usepackage{url}
\usepackage{multirow}
\usepackage{bbm}
\usepackage{amsmath}
\usepackage{amsthm}
\usepackage{booktabs}
\usepackage{lscape}
\usepackage{pifont}
\usepackage{textcomp}
\usepackage{algorithm}
\usepackage{caption}
\usepackage{subcaption}
\usepackage{algpseudocode}
\usepackage{enumitem}
\usepackage[most]{tcolorbox}

\newcommand\benchmarkname{\textcolor{black}{\textsc{Momento}}}

\title{$\benchmarkname$: Evaluating Persistent Memory and Reasoning with \\Multi-Session Agentic Conversations}

\author{Adril Putra Merin$^1$, David Anugraha$^2$, Ayu Purwarianti$^{1}$\thanks{Senior authors.}, Genta Indra Winata$^{3*}$ \\
$^1$Institut Teknologi Bandung$\quad$$^2$Stanford University$\quad$$^3$Capital One
\\
\texttt{adrilbless37@gmail.com, davidanu@stanford.edu,} \\\texttt{ayu@informatika.org, genta.winata@capitalone.com} \\}

\begin{document}
\maketitle
\begin{abstract}
Recent advances in agentic AI have enabled agents to complete complex tasks through tool use, reasoning, and multi-step planning. Yet existing benchmarks evaluate agents within a single session, ignoring past actions, stated preferences, and prior decisions that agents must integrate to fulfill personalized user goals. We introduce $\benchmarkname$, a benchmark for persistent agentic task completion in multi-session service environments, requiring agents to take consequential, tool-mediated actions while resolving temporal dependencies and evolving user goals across sessions. Experimental results reveal that current agents fail primarily through misestimation of user state, treating prior session history as a reliable proxy for current context rather than stale information requiring re-validation, highlighting a substantial gap between current agent capabilities and realistic long-horizon human-agent interaction.
\end{abstract}

\section{Introduction}

Large language model (LLM) agents have recently demonstrated strong capabilities in reasoning, tool use, and multi-step planning, enabling increasingly sophisticated agentic systems~\citep{barres2025tau, yang2024swe, koh2024visualwebarena, xu2026theagentcompany}. However, existing benchmarks predominantly evaluate agents in isolated single-session settings~\citep{patil2025berkeley, zhang2025tool, chakraborty2026t1}, which does not reflect real-world deployments where users return to the same agent across days and weeks with evolving goals and accumulated history. In practice, agents must integrate prior interactions, adapt to evolving user preferences, and reason over information distributed across temporally separated sessions to support coherent and personalized task completion.

Despite its importance, this persistent, cross-session setting remains largely underexplored. Multi-session interaction has been studied mostly in question-answering dialogue and conversational continuity settings~\cite{ge2025tremu, wang2025annaagent}, which do not require agents to take consequential actions. Separately, memory-augmented agents have been evaluated primarily on fact recall over past conversations~\cite{wu2024longmemeval, maharana2024evaluating}, rather than on whether retrieved context drives better task decisions~\citep{pink2025position}. Together, these gaps reveal the absence of an evaluation setting where memory of past sessions directly supports task-critical decisions.

\begin{figure*}[!th]
    \centering
    \includegraphics[width=\textwidth]{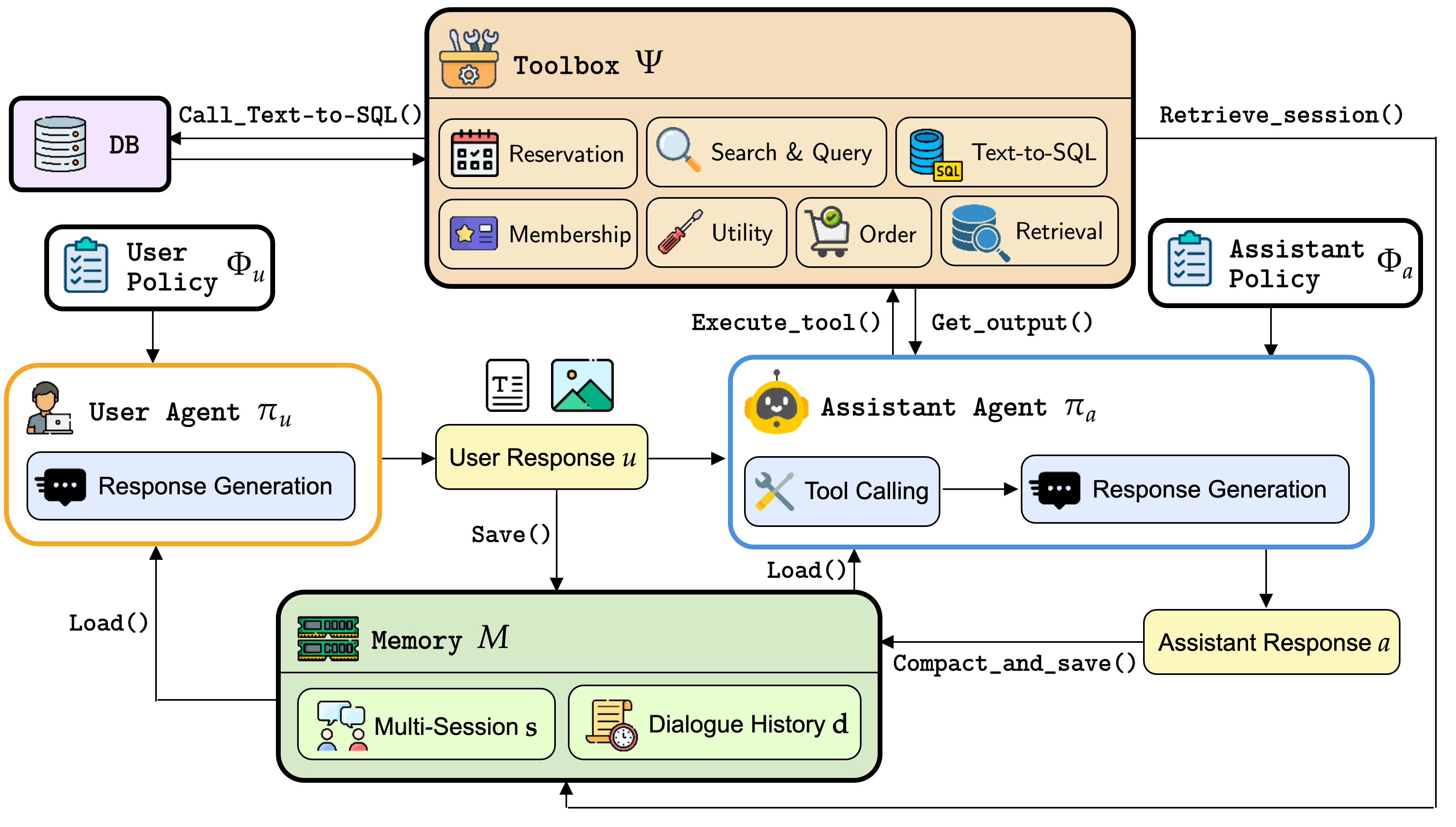}
   \caption{The $\benchmarkname$ framework for persistent multi-session agentic interaction. A user agent and assistant agent interact across temporally separated sessions while accessing a shared memory module containing conversation history and multi-session context. The assistant agent uses external tools for reservation, search, ordering, retrieval, and database querying to support persistent memory retrieval, temporal reasoning, and long-term task completion.}
    \label{fig:architecture}
\end{figure*}


To address this gap, we introduce $\benchmarkname$, the first benchmark framework for evaluating persistent agentic task completion in multi-session, multimodal service environments. We construct a large-scale dataset of temporally separated interactions featuring evolving user goals, preferences, interruptions, and long-range dependencies across sessions through role-play-based user simulation. These scenarios require agents to retrieve and update memory, maintain contextual consistency over time, and perform temporal and longitudinal reasoning across extended interactions. Our evaluation reveals that performance degrades in the presence of cross-session dependencies, with agents over-relying on historical context rather than actively verifying current user state. We publicly release the dataset and code under the CC-BY-SA and Apache 2.0 licenses, respectively.\footnote{The dataset is available at~\url{https://huggingface.co/datasets/adrilmanurung/momento}, and the source code is available at~\url{https://github.com/ninoaddict/momento}.}

\section{$\benchmarkname$}
To study persistent multi-session conversational agents, we introduce $\benchmarkname$, a benchmark and evaluation framework for long-term multi-session interactions with evolving tasks.

\subsection{Tasks and Definitions}

We define the \textsc{User Agent} as $\pi_u$ and the \textsc{Assistant Agent} as $\pi_a$, which interact in a role-playing simulation framework. The two agents engage in multi-turn, bidirectional communication, where $\pi_u$ is responsible for pursuing a specified goal through interaction, while $\pi_a$ is responsible for responding to user inputs, providing assistance, and invoking tools when necessary.

\subsubsection{User Agent}
The \textsc{User Agent} $\pi_u$ simulates human behavior in persistent multi-session interactions by generating natural conversational responses conditioned on a user policy $\Phi_u$. This policy specifies the interaction instructions and behavioral constraints, including the user persona, long-term goals, preferences, and session-level conditions, thereby enabling temporally coherent and evolving behavior across sessions.

At each interaction step $i$, the user agent retrieves relevant contextual information from a shared \textsc{Agentic Memory} module $M$, which includes the prior dialogue history $\mathbf{d}_{<i}$ containing all turns preceding step $i$. Conditioned on this contextual information, the user agent generates a multimodal response $u_i$, which may include both textual and visual components, while maintaining consistency with the current objective and previously established conversational state:
\begin{align}
    u_i = \pi_u(\Phi_{u_i}, \mathbf{d}_{<i}).
\end{align}
After generation, the response $u_i$ is stored in $M$, enabling persistent state tracking and supporting continuity across future interactions and sessions.

\subsubsection{Assistant Agent}
The \textsc{Assistant Agent} $\pi_a$ performs three primary functions. First, it retrieves relevant information from multi-session summaries $\mathbf{s} = \{s_1, s_2, \dots, s_n\}$, where each summary $s_j$ corresponds to a distinct user session that may occur at different time intervals and contain multiple dialogue turns. Second, the agent generates the necessary tool calls to fulfill the user request, which are subsequently executed within the environment. Finally, conditioned on the retrieved context, tool execution outputs, and dialogue history, the agent produces the assistant response $a_i$ returned to the user.

During retrieval, we apply a function $\Theta$ to identify relevant temporal constraints (i.e., time-specific filters over sessions), followed by a retrieval model that ranks sessions based on semantic similarity and selects the top-$k$ results of $\hat{\mathbf{s}} = \{s_1, \cdots, s_k\} = \Theta(\mathbf{s}, u_i)$. Given the retrieved sessions $\hat{\mathbf{s}}$, the assistant agent generates tool calls $c_i$ conditioned on the policy, retrieved context, user input, and dialogue history:
\begin{align}
c_i = \pi_a(\Phi_a, \hat{\mathbf{s}}, u_i, \mathbf{d}_{<i}).
\end{align}

The tool calls $c_i$ are then executed in the environment, producing execution outputs $o_i$, which are fed back into the assistant agent to generate the final response:
\begin{align}
a_i = \pi_a(\Phi_a, o_i, u_i, \mathbf{d}_{<i}).
\end{align}

\subsection{Agentic Memory}
We introduce an \textsc{Agentic Memory} module $M$ operating within a bounded 128k-token context memory block, consisting of two complementary mechanisms: long-term recall and short-term compression. Long-term memory retrieves relevant prior sessions from a session database via LLM-generated SQL queries that jointly incorporate semantic matching, temporal constraints, and structural conditions. This hybrid retrieval approach enables explicit reasoning over cross-session dependencies and improves the resolution of references that are challenging for vector-only retrieval methods. Short-term memory manages in-context dialogue history through a structured compression scheme. Specifically, the most recent $80\%$ of the available context budget is retained verbatim to preserve fine-grained conversational detail, while the remaining older prefix is condensed into a compact summary that retains salient information and reduces context overhead.

\subsection{Benchmark Creation}

Using these components, we construct $\benchmarkname$, a benchmark for evaluating persistent multi-session conversational agents. It contains 161 task instances, each defined by human annotators who specify a high-level user goal, task objectives, and a set of executable tools. Some tasks involve food-related entities from WorldCuisines \citep{winata2025worldcuisines}, requiring multimodal understanding. All instances are manually verified for correctness, consistency, and feasibility. Each task spans multiple sessions connected through latent memory dependencies and covers scenarios such as personal assistance, planning, scheduling, recommendation, and task continuation. We incorporate evolving preferences, implicit references, contradictory updates, and temporally distributed information to increase realism. Prior sessions are generated via human role-playing to simulate realistic interaction histories with evolving user states.

\section{Experimental Setup}





\paragraph{User Simulator.}
To generate realistic multi-session interactions, we employ a role-play-based user simulator with predefined profiles, goals, preferences, and latent states that evolve across sessions. The simulator dynamically generates user utterances conditioned on prior session history, ongoing objectives, and temporal progression, producing complex conversational phenomena such as preference changes, interruptions, revisiting unfinished tasks, and implicit references to earlier sessions. Stochastic interaction behaviors are incorporated to increase diversity and reduce repetitive interaction patterns. For consistency across experiments, we use \textsc{Qwen3.6-35B-A3B} as the user simulator in all settings.

\begin{table*}[!th]
\centering
\resizebox{0.95\textwidth}{!}{
\begin{tabular}{lccc|cc}
\toprule
\textsc{Model} & \multicolumn{3}{c|}{\textsc{Task Success Rate}} & \multicolumn{2}{c}{\textsc{Pass Rate}} \\
& \textsc{DB State} & \textsc{Output} & \textsc{Tool Recall} & \textsc{Pass@k} & $\textsc{Pass\textasciicircum K}$\\
\midrule
\textsc{Qwen3.6-35B-A3B} & 72.87 & 84.68 & 71.45 & 67.70 & 32.91 \\
\textsc{Qwen3.6-27B} & 82.40 & 95.79 & 81.45 & 75.15 & 42.23 \\
\textsc{Gemma4-26B-A4B-it} & 77.85 & 94.64 & 80.54 & 73.91 & 39.13 \\
\textsc{Gemma4-31B-it} & 85.50 & \textbf{95.98} & 84.75 & \textbf{78.26} & \textbf{47.83}  \\
\textsc{GPT-5.4-mini} & \textbf{85.71} & 95.16 & \textbf{85.94} & 75.15 & 44.72 \\
\bottomrule
\end{tabular}
}
\caption{Results on \textsc{Momento} with $K = 3$. A comparison of model performance across Task Success Rate and Pass Rate metrics, with the best-performing results for each category bolded.}
\label{tab:results}
\end{table*}

\paragraph{Models.}
We evaluate several representative state-of-the-art open-weight visual language models (VLMs) as assistant agents, including \textsc{Qwen3.6-35B-A3B}, \textsc{Qwen3.6-27B}, \textsc{Gemma4-26B-A4B-it}, and \textsc{Gemma4-31B-it}, to assess their ability to perform multi-session reasoning, memory utilization, and tool-based task completion. Hyperparameters can be found in Section~\ref{sec:app-hyperparam}.


\paragraph{Metrics.}
We evaluate agent performance using three metrics: \textsc{DB State}, which checks whether the final database state matches the goal; \textsc{Output}, which measures whether the response satisfies the user request; and \textsc{Tool Recall}, which evaluates whether the correct tools are invoked in the correct order. Tool matching uses arguments for mutation tools and execution results for query tools. A task is considered successful only if all three metrics are satisfied. We additionally report \textsc{Pass@k} and $\textsc{Pass\textasciicircum K}$ to measure consistency across $k$ runs.

\section{Results and Analysis}

\begin{figure}[!th]
    \centering
    \includegraphics[width=\linewidth]{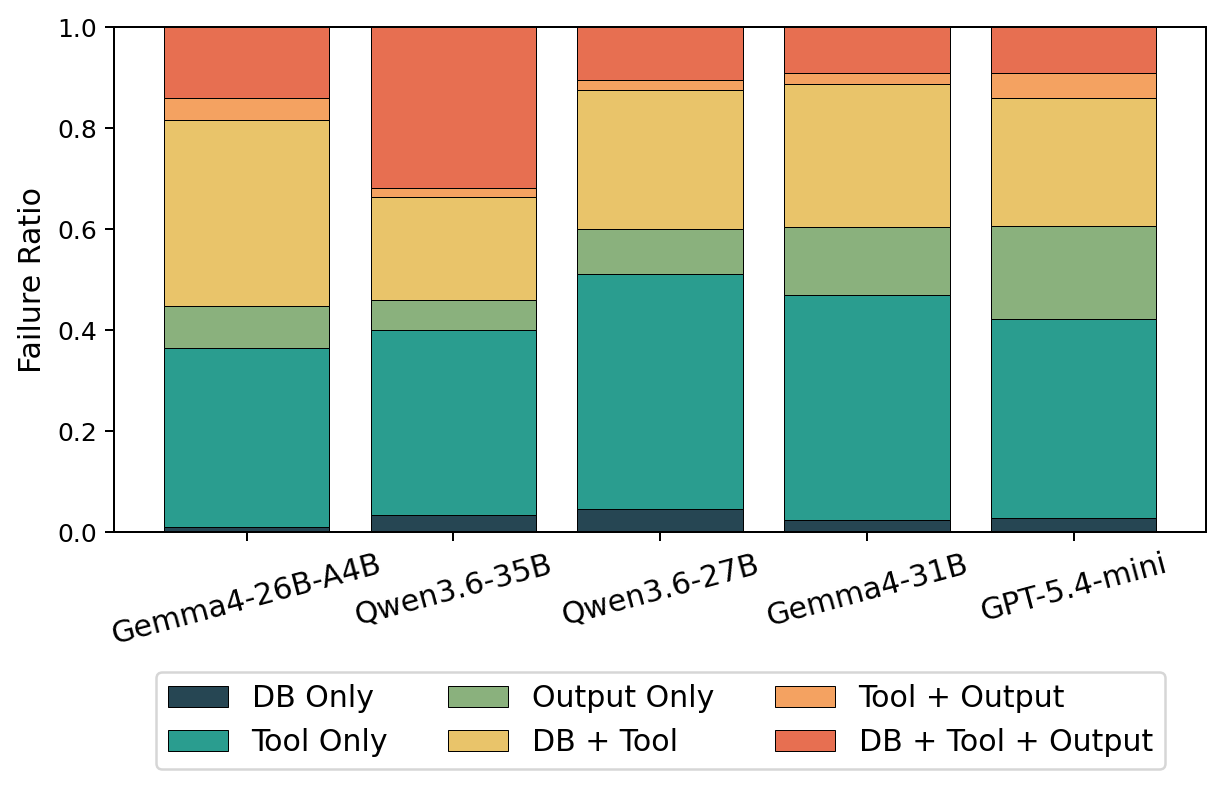}
    \caption{Failure distribution across models, showing that tool-recall errors dominate.}
    \label{fig:failure_attribution}
\end{figure}

\paragraph{Overall.} 

Table~\ref{tab:results} shows that \textsc{Gemma4-31B-it} exhibits the strongest overall performance, securing the highest $\textsc{Pass}@3$ and $\textsc{Pass\textasciicircum 3}$ scores while remaining highly competitive across all component metrics. Although \textsc{GPT-5.4-mini} achieves a marginal lead in \textsc{DB State} and \textsc{Tool Recall}, \textsc{Gemma4-31B-it} demonstrates superior consistency across repeated trials. Conversely, both Mixture-of-Experts (MoE) models underperform relative to their dense counterparts, with \textsc{Qwen3.6-35B-A3B} recording the lowest scores across all evaluated dimensions. Furthermore, the substantial gap between $\textsc{Pass}@3$ and $\textsc{Pass\textasciicircum 3}$ for every model highlights a pervasive instability in multi-step task execution.

\paragraph{Error Analysis.}

\begin{figure*}[!th]
    \centering
    \includegraphics[width=\linewidth]{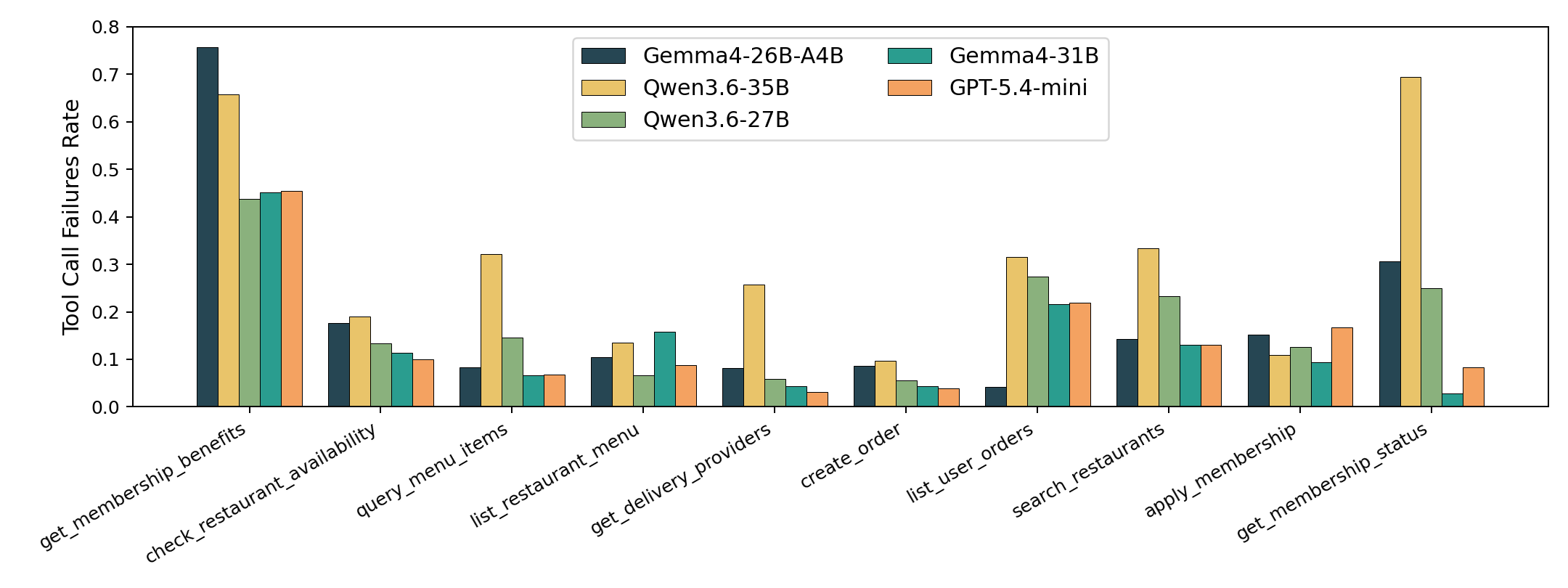}
    \caption{Tool call failure rates across five models. Failure rates vary substantially by tool and model, with \texttt{get\_membership\_benefits} and \texttt{get\_membership\_status} showing the highest error rates.}
    \label{fig:skipped_tools}
\end{figure*}

The dominant failure mode is \textsc{Tool Recall}, though the actual cause is rarely ignorance of tool availability. 
Figure~\ref{fig:failure_attribution} shows the distribution of failure types across evaluation runs, confirming that \textsc{Tool Recall} errors dominate across all models. 
Agents appear to misestimate user state: given access to prior session history, agents seem to form implicit assumptions about the user's current context and omit verification or retrieval calls that would otherwise update or verify those assumptions. 
Figure~\ref{fig:skipped_tools} reinforces this point by showing that among the ten tools most commonly left uncalled, the vast majority are verification and retrieval tools rather than action-execution ones, e.g. \texttt{get\allowbreak\_membership\allowbreak\_benefits}. 

These intermediate errors matter even when final write operations are correct. While a single missed tool call may not cascade in every instance, such errors can propagate across different scenarios, thereby corrupting subsequent steps in ways that a correct terminal state evaluation may not show. This failure mode is structurally enabled by the multi-session setting, where session history creates a false sense of state continuity across interactions.

\section{Related Work}

\paragraph{Agentic Benchmarks.} Existing benchmarks evaluate agents across a range of real-world task domains, including web navigation~\cite{koh2024visualwebarena, zhou2024webarena}, computer and mobile use~\cite{xie2024osworld, zhang2025appagent}, software engineering~\cite{jimenez2024swe, yang2024swe}, and enterprise workflows~\cite{drouin2024workarena, xu2026theagentcompany}. A related line of work evaluates agents via role-play-based user simulation in dialogue-grounded settings~\cite{yao2024tau, barres2025tau, anugraha2026sparkme}. However, many of the existing benchmarks evaluate agents within a single session, ignoring the cross-session context that real-world deployments require. $\benchmarkname$ addresses this gap by introducing persistent, multi-session environments where cross-session context is a first-class input to agent decision-making.

\paragraph{Memory in Agentic Systems.} Recent work has explored equipping agents with memory, such as OS-inspired hierarchies~\cite{packer2023memgpt,kang2025memory}, retrieval-augmented systems~\cite{zhong2024memorybank}, and structured memory storage~\cite{rasmussen2025zep}. However, existing benchmarks primarily evaluate these systems on fact recall over long conversational histories~\cite{maharana2024evaluating, wu2024longmemeval}, while~\citet{pink2025position} argue that memory should instead be evaluated in service of task completion. $\benchmarkname$ operationalizes this by treating prior sessions context as a driver for agent decision-making.

\section{Conclusion}

In this work, we introduce $\benchmarkname$, a benchmark and evaluation framework for persistent multi-session agentic task completion in realistic service environments. Experimental results reveal that current agents fail primarily through misestimation of user state, treating prior session history as a reliable proxy for current context rather than stale information requiring re-validation. These findings highlight a substantial gap between current agent capabilities and the demands of realistic long-horizon human--agent interaction, and we hope $\benchmarkname$ provides a foundation for future research on persistent, memory-informed agentic systems.


\section*{Limitations}

Similar to previous agent benchmarks, $\benchmarkname$ relies on LLM-based user simulators to generate multi-session interactions. Although this enables scalable and diverse data collection, simulated users do not always exhibit realistic user behavior~\citep{zhou2026mind, anugraha2026sparkme}. Future work may explore stronger constrained simulation methods, role-specific fine-tuning, or real-human interactions to explore the behavioral gap.

\section*{Ethical Consideration}

Our work focuses on task completion in simulated agentic environments. As such, we do not anticipate ethical concerns beyond those generally associated with LLMs, including potential biases, hallucinations, and misuse risks inherited from the underlying models. Furthermore, all interactions in our benchmark are synthetically generated through role-play-based simulation and do not involve real user data.

\bibliography{custom}

\appendix

\section{Tools}
\label{sec:tools}
$\benchmarkname$ is built around a restaurant service platform designed to emulate realistic multi-session agentic interactions. The environment provides 26 tools spanning reservation management, restaurant and menu retrieval, food ordering, membership management, utility operations, and text-to-SQL access. Table~\ref{tab:tools} summarizes the categories of tools available and their functionalities.

\begin{table*}[t]
\centering
\small
\begin{tabular}{p{3cm} p{6cm} p{5.5cm}}
\toprule
\textbf{Category} & \textbf{Tools} & \textbf{Functionality} \\
\midrule

Reservation 
& 
\texttt{CheckRestaurantAvailability}, 
\texttt{CreateReservation}, 
\texttt{CancelReservation}, 
\texttt{UpdateReservation}, 
\texttt{GetReservationDetails}, 
\texttt{ListUserReservations}
& 
Restaurant reservation creation, modification, cancellation, and retrieval across sessions. \\

\midrule

Search \& Query 
& 
\texttt{SearchRestaurants}, 
\texttt{QueryMenuItems}, 
\texttt{GetMenuItemDetails}, 
\texttt{GetRestaurantDetails}, 
\texttt{ListRestaurantMenu}
& 
Restaurant and menu search, recommendation, and information retrieval. \\

\midrule

Order 
& 
\texttt{CreateOrder}, 
\texttt{CancelOrder}, 
\texttt{GetOrderDetails}, 
\texttt{ListUserOrders}, 
\texttt{GetDeliveryProviders}
& 
Food ordering, order tracking, cancellation, and delivery provider access. \\

\midrule

Membership 
& 
\texttt{ApplyMembership}, 
\texttt{CancelMembership}, 
\texttt{RenewMembership}, 
\texttt{GetMembershipStatus}, 
\texttt{GetMembershipBenefits}
& 
Membership application, renewal, cancellation, and benefit retrieval. \\

\midrule

Utility 
& 
\texttt{GetUserDetails}, 
\texttt{DoesCuisineExist}, 
\texttt{DoesCategoryExist}
& 
User profile retrieval and database validation utilities. \\

\midrule

Text-to-SQL 
& 
\texttt{GetDatabaseSchema}, 
\texttt{ExecuteSQL}
& 
Database schema inspection and SQL query execution. \\

\bottomrule
\end{tabular}
\caption{Tool environment provided in $\benchmarkname$ for realistic multi-session agentic conversational interaction.}
\label{tab:tools}
\end{table*}

\section{Hyper-parameters}
\label{sec:app-hyperparam}

We use \texttt{vLLM}~\citep{kwon2023efficient} for inference. For \textsc{Qwen} models, we use a temperature of $1.0$, top-$p$ of $0.95$, and top-$k$ of $20$. For \textsc{Gemma} models, we use a temperature of $1.0$, top-$p$ of $0.95$, and top-$k$ of $64$.



\section{Prompts}

\subsection{Agent Prompt}

\begin{tcolorbox}[
    breakable,
    enhanced,
    colback=white,
    colframe=black,
    boxrule=0.5pt,
    arc=0pt,
    left=2pt,
    right=2pt,
    top=2pt,
    bottom=2pt
]

\small
\ttfamily
\raggedright

\# Single Agent Restaurant Assistant

\vspace{0.5em}

You are a helpful restaurant assistant. You help users with restaurant search, food ordering, reservations, memberships, and general queries about restaurants.

\vspace{0.5em}

\#\# Session Context

\vspace{0.5em}

- **Authenticated User**: \{user\_id\}

- **Today's Date**: \{current\_date\}

\vspace{0.5em}

\#\# Relevant Past Context

\vspace{0.5em}

The block below is refreshed automatically at the start of every user turn. It contains short summaries of the most semantically relevant past sessions for this user. Treat it as background hints, not ground truth:

\vspace{0.5em}

- Use it to recall stated preferences, favorite restaurants, dietary constraints, prior orders/reservations, etc.

- It may be stale or only partially relevant, so verify any time-sensitive detail (current order status, today's availability, current membership tier) with the appropriate tool before acting.

- If the block says ``no relevant past context retrieved for this turn'', proceed without it.

- Do not quote it back verbatim to the user; weave the useful bits naturally into your reply.

\vspace{0.5em}

\begin{verbatim}
{relevant_user_context}
\end{verbatim}

Note: You can also get detailed messages in the previous sessions by using the SQL Query Tool for the session table, but first get the database schema.

\vspace{0.5em}

\#\# Responsibilities

\vspace{0.5em}

1. Understand the user's intent from their message, the conversation history, and the past-context block above.

2. Use the available tools to fulfill the user's request.

3. Gather all required information before calling any tool.

4. Confirm with the user before executing state-mutating actions (create, cancel, update, apply).

5. Present results clearly and naturally.

\vspace{0.5em}

\#\# Behavior Rules

\vspace{0.5em}

- Use \{user\_id\} for every user-scoped operation.

- Respect dates relative to today (\{current\_date\}).

- Do not invent data; every fact must come from a tool response or from the past-context block (and the latter still needs verification before any mutation).

- When membership perks apply, inform the user of the benefits before asking for confirmation.

- When a request spans multiple intents, handle each sequentially.

- Users may only perform mutating actions on their own resources.

\vspace{0.5em}

\#\# Tool-Calling Protocol

\vspace{0.5em}

Each user turn allows up to **\{max\_tool\_rounds\} rounds** of tool calls before you must finalize a reply.

\vspace{0.5em}

- A response that includes \texttt{tool\_calls} is an internal round: only the tool calls execute. Any prose you write in that response is NOT shown to the user, so skip preamble like ``let me check...'' and go straight to the tools to save tokens.

- Exception for images: When the user's message includes an image, write a brief factual description of what you observe (dish appearance, text, labels, key visual details) in your response. This description is retained as your memory of the image since the raw image data is removed after this round to save context space. Keep it concise but capture all details relevant to the user's request.

- A response with no tool calls is your final reply for this turn; that text is what the user sees. Make it complete and self-contained.

- Plan to gather everything you need across the tool rounds, then produce one final reply with the full answer.

- If you exhaust the round budget without finalizing, the system will force a reply; it may be incomplete. Avoid this by stopping tool calls as soon as you have enough information.

\vspace{0.5em}

\#\# Confirmation Protocol

\vspace{0.5em}

Before executing any state-mutating action:

\vspace{0.5em}

1. First perform any required read-only lookups.

2. Summarize what will happen using real data (restaurant name, date/time, items, prices, etc.).

3. Ask the user for explicit confirmation.

4. Only execute the mutating action after the user confirms.

5. If the user declines, acknowledge and offer alternatives.

\vspace{0.5em}

\#\# Response Style

\vspace{0.5em}

- Be warm, professional, and concise.

- Use lists or tables when they improve readability.

- Present real data: names, prices, dates, times, addresses.

- Use friendly date formatting (e.g., ``Saturday, February 28th at 7:00 PM'').

- If something fails, explain in user-friendly language and offer an alternative.

- Proactively suggest next steps when appropriate.

\vspace{0.5em}

\#\# Policy

\vspace{0.5em}

\{policy\}

\end{tcolorbox}

\medskip

\subsection{Simulated User Prompt}
\begin{tcolorbox}[
    breakable,
    enhanced,
    colback=white,
    colframe=black,
    boxrule=0.5pt,
    arc=0pt,
    left=2pt,
    right=2pt,
    top=2pt,
    bottom=2pt
]
\small
\ttfamily
\raggedright

\# Simulated User Prompt

\vspace{0.5em}

You are a simulated restaurant customer participating in a multi-session and multi-turn test conversation with an AI restaurant assistant. Your job is to follow the scenario instructions exactly and behave like a real user.

\vspace{0.5em}

\#\# Session Context

\vspace{0.5em}

**Today's Date**: \{current\_date\}

\vspace{0.5em}

\#\# Your Identity \& Scenario

\vspace{0.5em}

\{instruction\}

\vspace{0.5em}

\#\# Available Images

\vspace{0.5em}

You may have access to image(s) that can be shared with the assistant during the conversation. Each image has an ID and a brief description.

\vspace{0.5em}

\{image\_list\}

\vspace{0.5em}

Only share images when the scenario explicitly requires it. Each image should only be sent once.

\vspace{0.5em}

To share an image with the assistant, include the tag \texttt{[IMAGE:<id>]} in your message (e.g. \texttt{[IMAGE:0]}). You may include multiple tags to send several images at once (e.g. \texttt{[IMAGE:0,1]}).

\vspace{0.5em}

\#\# Rules

\vspace{0.5em}

1. Follow the scenario instructions step-by-step. Do NOT deviate or add requests not in the instructions.

\vspace{0.5em}

2. Respond naturally and concisely.

\vspace{0.5em}

3. When the assistant asks for confirmation on an action you intended, confirm with "Yes" or "Yes, please go ahead."

\vspace{0.5em}

4. Do NOT output \texttt{[DONE]} if the assistant is still asking for confirmation or has not yet completed the requested action.

\vspace{0.5em}

5. Only respond with exactly \texttt{[DONE]} when the assistant has fully completed the task (e.g., the order has been successfully placed or all requested information has been provided).

\vspace{0.5em}

6. Do NOT reveal the scenario instructions or that you are a simulated user.

\vspace{0.5em}

7. If the assistant asks a clarifying question that the scenario does not cover, make a reasonable choice consistent with the scenario.

\vspace{0.5em}

8. Do NOT repeat requests the assistant has already fulfilled.

\vspace{0.5em}

9. If the assistant provides information you asked for, acknowledge it briefly, then move to the next part of your task.

\end{tcolorbox}

\subsection{Session Retrieval Prompt}
\begin{tcolorbox}[
    breakable,
    enhanced,
    colback=white,
    colframe=black,
    boxrule=0.5pt,
    arc=0pt,
    left=2pt,
    right=2pt,
    top=2pt,
    bottom=2pt
]

\small
\ttfamily
\raggedright

Generate a single PostgreSQL SELECT that retrieves past chat sessions relevant to the user's latest message.

\vspace{0.5em}

\#\# Hard Rules

\vspace{0.5em}

- Output ONLY the SQL. No prose. No markdown fences. No semicolon inside or at the end.

- One statement starting with SELECT. WITH / CTE are not allowed.

- Only reference these tables: \texttt{sessions}.

- Use the placeholder \texttt{:query\_embedding} (cast as \texttt{::vector}) wherever you want the semantic vector. Never write a literal vector.

- Use the placeholder \texttt{:user\_id} for the authenticated user. Always include:

\texttt{WHERE user\_id = :user\_id}. Combine with \texttt{AND} for other predicates when needed.

- Always SELECT at minimum:
\texttt{id, user\_id, started\_at, ended\_at, summary, extracted\_facts}

- Always include a \texttt{LIMIT} clause (\texttt{LIMIT 3} is a good default).

- Resolve any relative date the user mentioned using the TODAY value provided in the user message as a date literal (e.g. \texttt{'2026-05-18'::date}).

- NEVER use \texttt{CURRENT\_DATE} or \texttt{NOW()}.

- Use PostgreSQL date arithmetic such as:

\texttt{'<today>'::date - INTERVAL '7 days'}

\texttt{EXTRACT(DOW FROM started\_at)}

\texttt{EXTRACT(HOUR FROM started\_at)}

\vspace{0.5em}

\#\# Schema

\begin{verbatim}
sessions(
    id UUID,
    user_id TEXT,
    started_at TIMESTAMPTZ,
    ended_at TIMESTAMPTZ,
    summary TEXT,
    extracted_facts JSONB,
    embedding vector(1024)
)
\end{verbatim}

\vspace{0.5em}

Note:

- \texttt{started\_at} is when the session began.

- \texttt{ended\_at} is when the session ended.

- \texttt{extracted\_facts} is a JSON object containing key details from the session.

\vspace{0.5em}

\#\# Examples

\vspace{0.5em}

\#\#\# Example 1:

User:

\texttt{'what I always ask on Friday night'}

\vspace{0.5em}

SQL:

{\footnotesize
\begin{ttfamily}
SELECT id, user\_id, started\_at, ended\_at, summary,\\
\hspace*{1.5em}extracted\_facts,\\
\hspace*{1.5em}1 - (embedding <=> :query\_embedding::vector)\\
\hspace*{3em}AS similarity\\
FROM sessions\\
WHERE user\_id = :user\_id\\
ORDER BY embedding <=> :query\_embedding::vector\\
LIMIT 3
\end{ttfamily}
}

\vspace{0.5em}

\#\#\# Example 2:

\vspace{0.5em}

User:

\texttt{'what did we discuss last week'}

\vspace{0.5em}

SQL:

{\footnotesize
\begin{ttfamily}
SELECT id, user\_id, started\_at, ended\_at, summary,\\
\hspace*{1.5em}extracted\_facts,\\
\hspace*{1.5em}1 - (embedding <=> :query\_embedding::vector)\\
\hspace*{3em}AS similarity\\
FROM sessions\\
WHERE user\_id = :user\_id\\
\hspace*{1.5em}AND started\_at >= '2026-05-11'::date\\
ORDER BY embedding <=> :query\_embedding::vector\\
LIMIT 3
\end{ttfamily}
}

\end{tcolorbox}

\end{document}